\pdfoutput=1
\documentclass[journal,twoside,web]{ieeecolor}
\usepackage{graphicx}
\usepackage{xstring}

\let\ORIincludegraphics\includegraphics
\renewcommand{\includegraphics}[2][]{%
  \IfSubStr{#2}{\logoname.eps}%
    {}%
    {\ORIincludegraphics[#1]{#2}}%
}
\usepackage{generic}
\usepackage{cite}
\usepackage{amsmath,amssymb,amsfonts}
\usepackage{algorithmic}
\usepackage{graphicx}
\usepackage{algorithm,algorithmic}
\usepackage{hyperref}
\hypersetup{hidelinks=true}
\usepackage{textcomp}
\usepackage{subfigure}
\usepackage{graphicx}

\def\BibTeX{{\rm B\kern-.05em{\sc i\kern-.025em b}\kern-.08em
    T\kern-.1667em\lower.7ex\hbox{E}\kern-.125emX}}
\begin{document}
\title{Attention-based Pin Site Image Classification in Orthopaedic Patients with External Fixators}
\author{Yubo Wang, Marie Fridberg, Anirejuoritse Bafor, Ole Rahbek, Christopher Iobst, Søren Vedding Kold, and Ming Shen,~\IEEEmembership{Senior Member,~IEEE}
\thanks{Manuscript received 29 September, 2023; revised XXX XX, XXXX. This work was supported in part by the China Scholarship Council.(Corresponding authors: Ming Shen; Søren Vedding Kold.)}
\thanks{Yubo Wang, Ming Shen are with the Department of the Electronic Systems, Aalborg University, 9220 Aalborg, Denmark (e-mail: yubow@es.aau.dk; mish@es.aau.dk).}
\thanks{Marie Fridberg, Søren Vedding Kold, Ole Rahbek are with the Department of Orthopaedic Surgery, Aalborg University Hospital, 9000 Aalborg, Denmark (e-mail: mfridberg@hotmail.com; sovk@rn.dk; o.rahbek@rn.dk).}
\thanks{Anirejuoritse Bafor, Christopher Iobst are with Center for Limb Lengthening and Reconstruction, Nationwide Children's Hospital, Columbus, Ohio 43205, USA (e-mail: Anirejuoritse.Bafor@nationwidechildrens.org; Christopher.Iobst@nationwidechildrens.org).}}

\maketitle
\begin{abstract}
Pin sites represent the interface where a metal pin or wire from the external environment passes through the skin into the internal environment of the limb. These pins or wires connect an external fixator to the bone to stabilize the bone segments in a patient with trauma or deformity. Because these pin sites represent an opportunity for external skin flora to enter the internal environment of the limb, infections of the pin site are common. These pin site infections are painful, annoying, and cause increased morbidity to the patients. Improving the identification and management of pin site infections would greatly enhance the patient experience when external fixators are used. For this, this paper collects and produces a dataset on pin sites wound infections and proposes a deep learning (DL) method to classify pin sites images based on their appearance: Group A displayed signs of inflammation or infection, while Group B showed no evident complications. Unlike studies that primarily focus on open wounds, our research includes potential interventions at the metal pin/skin interface. Our attention-based deep learning model addresses this complexity by emphasizing relevant regions and minimizing distractions from the pins. Moreover, we introduce an Efficient Redundant Reconstruction Convolution (ERRC) method to enhance the richness of feature maps while reducing the number of parameters. Our model outperforms baseline methods with an AUC of 0.975 and an F1-score of 0.927, requiring only 5.77 M parameters. These results highlight the potential of DL in differentiating pin sites only based on visual signs of infection, aligning with healthcare professional assessments, while further validation with more data remains essential.
\end{abstract}


\section{Introduction}
\label{sec:introduction}
\IEEEPARstart{A}{I} has come a long way and has shown great potential in many fields. Among them, image classification is a very active research direction in the field of computer vision, machine learning, and deep learning. It is widely used in many fields such as face recognition in security field\cite{liu2015targeting}\cite{zhao2019object}, anomaly detection\cite{pang2021deep}\cite{di2021pixel}, and medical image analysis\cite{song2021survey}\cite{litjens2017survey}\cite{cheng2022resganet} such as thyroid nodule detection\cite{10091892}, tissue segmentation\cite{scebba2022detect}, and coronavirus disease\cite{zhao2021deep}, etc. However, in terms of its application in orthopaedics, especially in scenarios involving the use of metal pins, its application is limited. This work aims to explore its potential in this scenario.

The external fixator (EF) is a common method used by orthopaedic surgeons to preserve, correct, and reconstruct the lower limb after a fracture or deformity. It consists of an external frame and pins that pass through the skin and are screwed into the bone tissue inside the limb. The pin site represents the area where the metal pin is passing through the skin from the external environment to the internal environment of the limb. Previous research indicates that up to 50\% of patients may experience infections ranging from mild to severe at pin sites\cite{potgieter2020complications}\cite{bue2021prospective}. Infections at surgical pin sites can lead to the instability of the EF, delayed bone healing, and potentially lead to serious complications such as deep infection requiring removal of the pin and intravenous antibiotics\cite{ceroni2016prevention}.

\begin{figure*}[!t]
\centering
\includegraphics[width=6.5in]{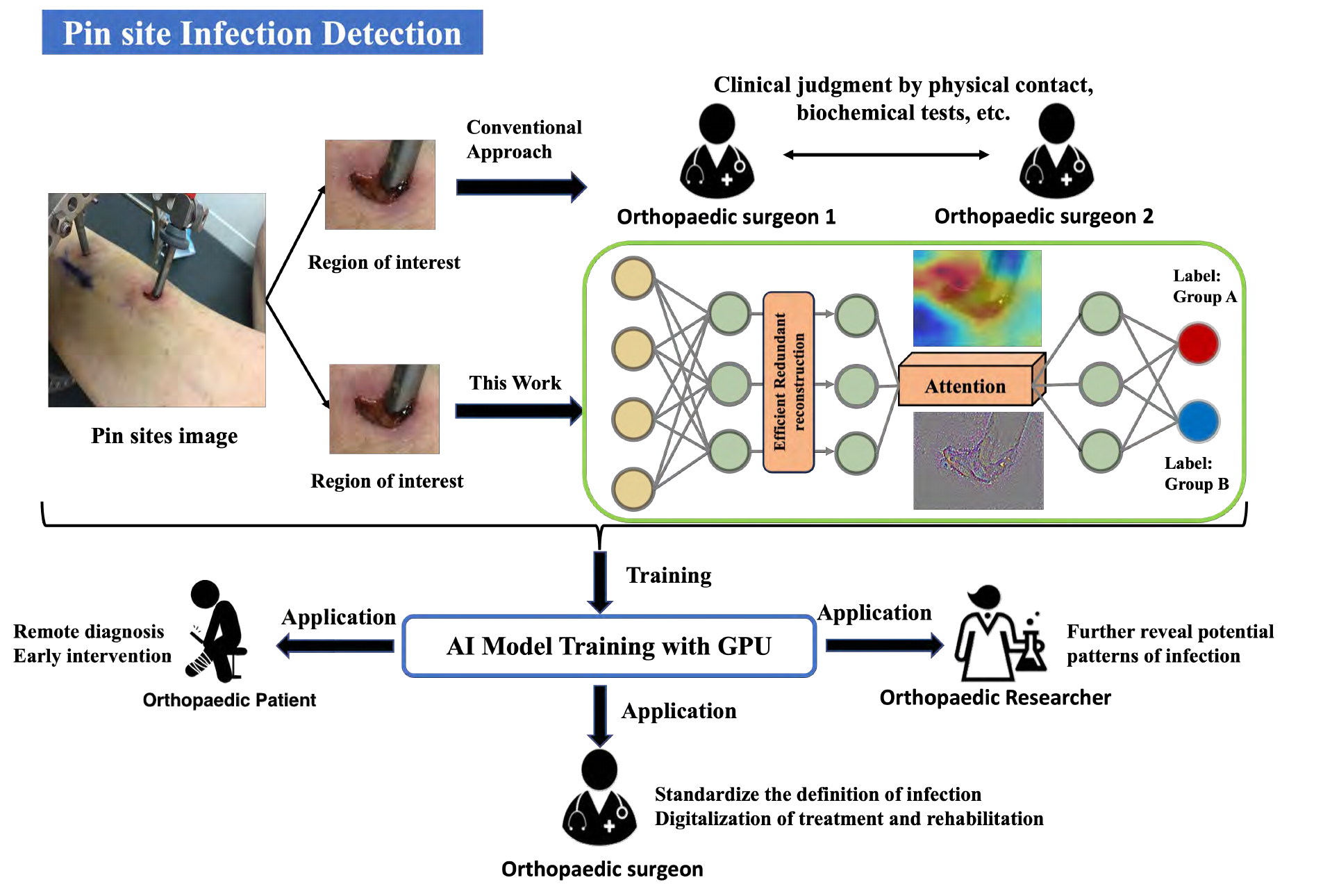}
\caption{Traditional and deep learning-based diagnosis workflow of wound status in lower extremity immobilization surgery. Doctors make a judgement on the inflammation/infection by the appearance of the wound. The deep learning method, which uses an attention mechanism, categorizes the skin tissues surrounding the wound by extracting their features.}
\centering
\label{fig_1}
\end{figure*}

Therefore, the assessment of the status of pin sites is crucial to treatment. However, pin site infections are typically determined by patient feedback, physical examination\cite{lethaby2013pin}, evaluation of radiographs and sometimes pin site fluid culture and blood work. Despite extensive research on the pathophysiologic mechanisms of pin site infections, accurate prediction of such infections remains a clinical challenge\cite{annadatha2022preparing}\cite{checketts2000orthofix}\cite{fragomen2017prophylactic}\cite{rahbek2021intrarater}\cite{frank2024practical}. Deep neural networks offer a promising solution, as they are highly capable of extracting information and learning valid representations of data from training data and labels without human or expert intervention. This capability makes them a valuable tool for automatically and remotely distinguishing between different types of pin site statuses. In previous studies, Yadav et al.\cite{yadav2019feature} used SVM to classify burn wound images based on color features. Abubakar et al.\cite{abubakar2020can} proposed a machine learning based approach to differentiate between burn wounds and pressure sores. The approach involved pre-training a deep architecture consisting of VGG-face, ResNet101, and ResNet152 to extract features, which were then fed into an SVM classifier for the classification task. Goyal et al.\cite{goyal2020recognition} proposed a deep learning-based classification method to predict the presence of infection or ischemia in Diabetic Foot Ulcers (DFUs). Rozo et al.\cite{10103156} proposed a U-net architecture with a VGG-based encoder for burn assessment on a small burn dataset. Nilsson et al.\cite{nilsson2018classification} used a CNN-based (Convolution Neural Network) approach to classify images of venous ulcers, utilizing VGG-19 network to classify images into venous or non-venous categories. In these classification problems, the deep learning-based approach outperformed traditional classifiers, demonstrating the potential benefits of using deep neural networks for remote wound condition assessment, such as infection. These benefits include improved accuracy, reduced healthcare costs, and increased convenience for patients. 

However, in convolutional neural networks such as VGG, ResNet, etc., the generated feature maps contain many similar features which are called redundant information. Kai et al.\cite{wang2021model} propose to reproduce this feature similarity pair with GhostNet to improve the classification ability of the model. This enhancement becomes particularly relevant in the case of large datasets like ImageNet and CIFAR-10, where the dataset is rich in features. In such datasets, similar features often play a significant role in contributing to the classification results, necessitating certain operations to amplify these similarities. For smaller datasets with limited features, avoiding the generation of similar feature pairs is the key to improving the classification results of the network. Therefore, the focus should be on generating more comprehensive features and identifying those that have the most significant impact on the classification results.

This work builds on the growing trend of using ubiquitous mobile devices as a healthcare decision-support tool. It offers the opportunity for smartphone-based cameras to detect and monitor the condition of surgical pin sites in home situations. Several organizations have developed fracture monitors to monitor wound healing, notably the Swiss AO Foundation, which developed an implantable biofeedback sensor device\cite{AO_Fracture_Monitor} for continuous monitoring of fracture healing status and patient activity. However, invasive monitors are not the optimal option for fracture wounds, which require a long recovery time. Other solutions require extra hardware, such as Swift Medical, which is looking to develop artificial intelligence-assisted surgical wound analysis by loading additional hyper-spectral sensor on the smartphone camera. This additional sensor enables long-wave infrared, near-infrared, ultraviolet, and enhanced RGB imaging to address wound care and management. Although this method is non-invasive, hyperspectral sensors are costly, which makes the technology expensive.

This study aims to apply deep learning to the analysis of images from pin sites to distinguish between different types of statuses based on their appearance. Additionally, it seeks to address the challenge of classifying pin site images in the presence of pin interference caused by the presence of metal pins. Figure. \ref{fig_1} shows a graphical workflow for wound type diagnosis in lower limb orthopaedic surgery based on traditional and deep learning methods. By developing a more accurate and efficient approach to wound classification, this work has the potential to significantly improve wound care and reduce healthcare costs.

\begin{figure*}[!t]
\centering
\includegraphics[width=6.in]{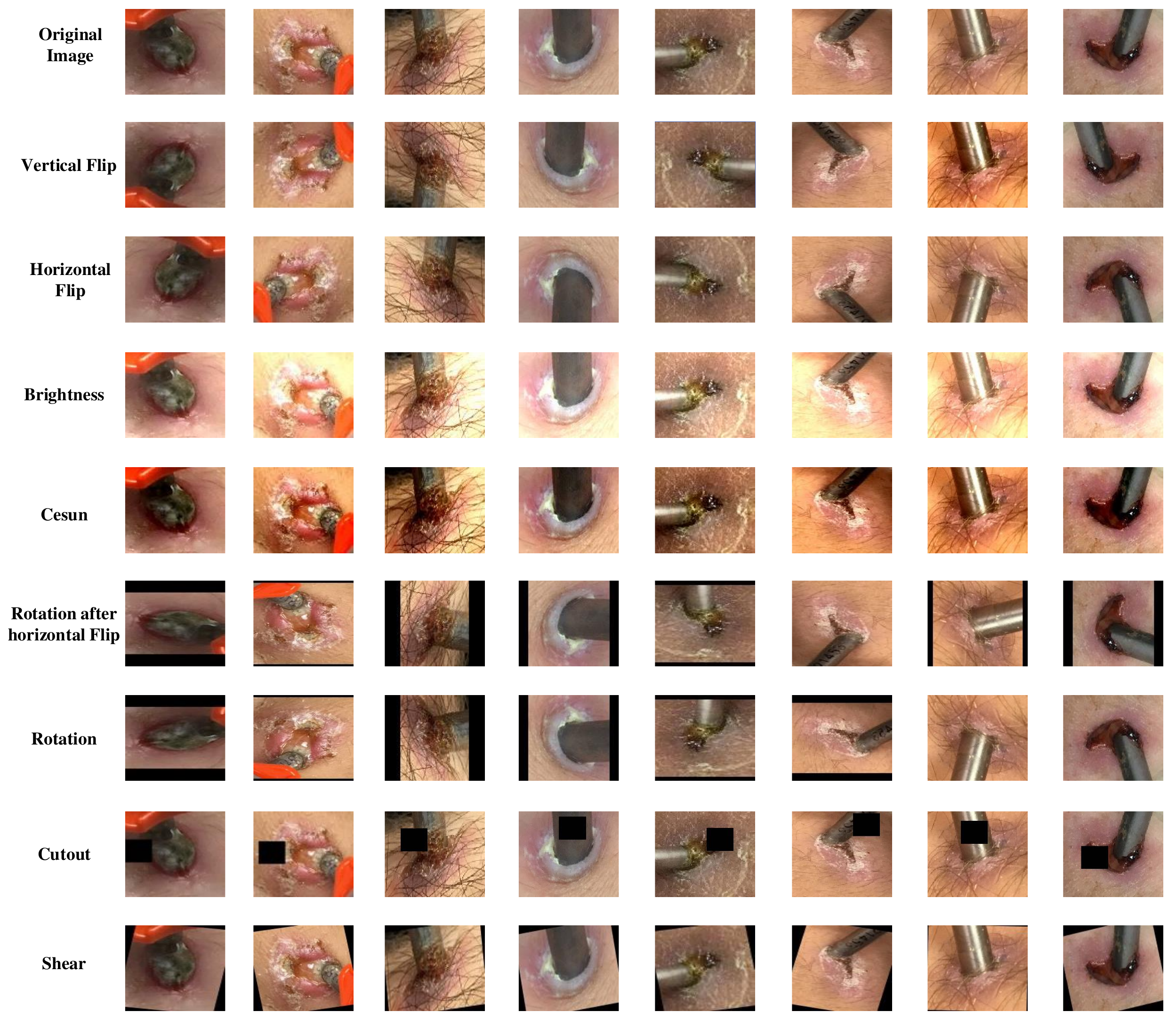}
\caption{Augmentation techniques applied to images, including flipping up down, changing brightness, changing contrast, flipping left right, rotating, and cutout.}
\centering
\label{fig_2}
\end{figure*}

\begin{figure*}[!t]
\centering
\includegraphics[width=6.3in, height=4in]{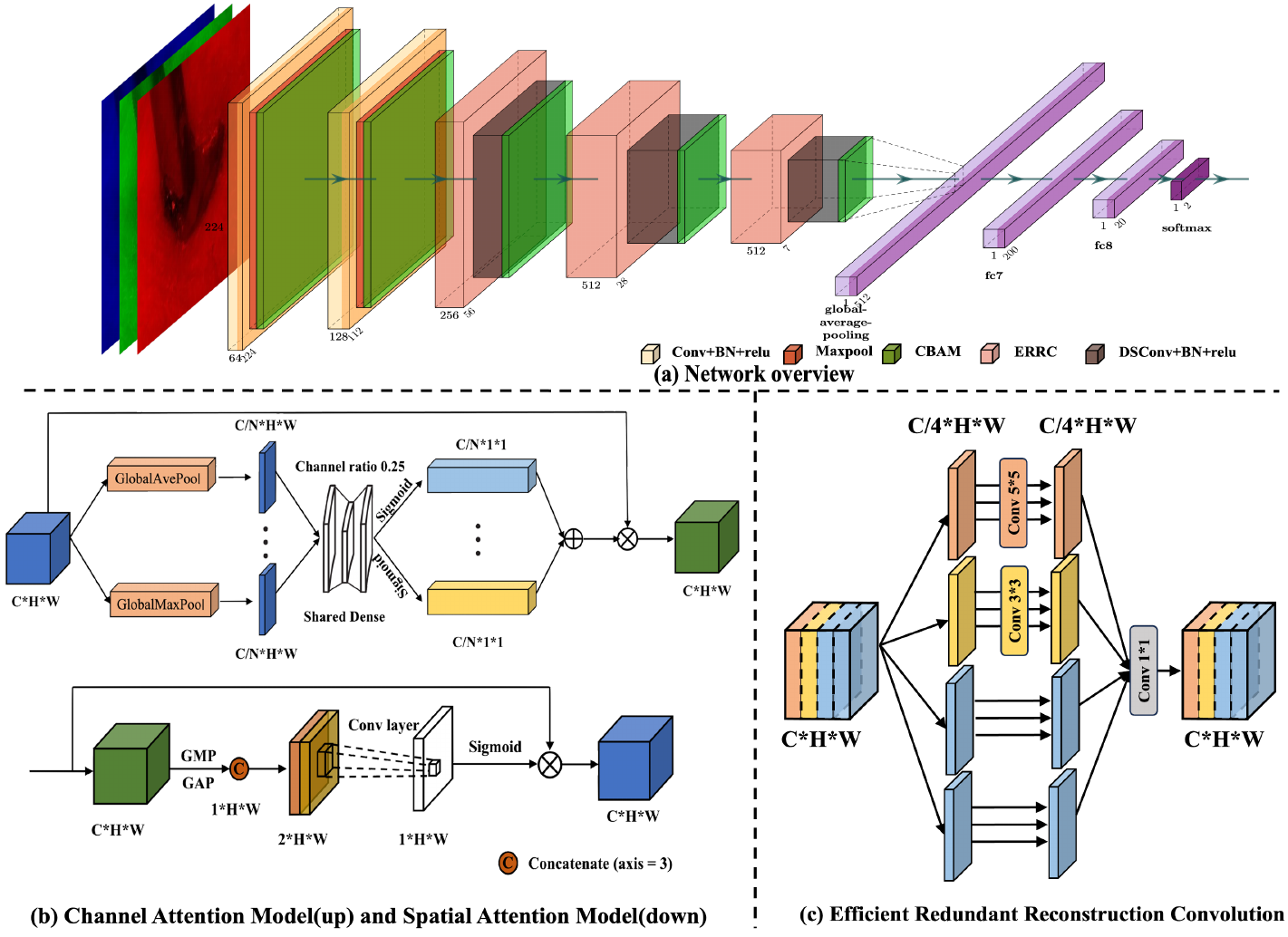}
\caption{(a) The architecture of our proposed network; (b) CBAM: the details of the channel and spatial attention modules; (c) Details of efficient redundant reconstruction convolution blocks, enriching the feature map by applying different convolutional kernels to different groups of channels.}
\centering
\label{fig_3}
\end{figure*}

\section{Method}
In this section, we first present the details of the NCH (Nationwide Children's Hospital) dataset, which we collected and created to facilitate the training and evaluation of our network. Next, we describe the network architecture we constructed for distinguishing different states of the pin sites from this dataset. We also provide insights into the architecture of various blocks used in the network. The following section will delve into the details and implementation of the dataset.

\subsection{Datasets and Protocols}
The image dataset was collected at the Nationwide Children's Hospital, Ohio, USA, with the primary goal of assisting in the analysis and diagnosis of patients undergoing treatment for fractures. The Research Institute at Nationwide Children's Hospital, through its IRB/ethics board, conducted this study and was responsible for the exemption determination. This comprehensive dataset comprises a total of 1554 high-resolution RGB images, each with a resolution of 3264 x 2448 pixels. The images were captured with an iPad 5th generation with an exposure time of 1/30 sec, an ISO speed rating setting of 80, and an aperture value is f/2.4. We subsequently pre-processed the dataset by removing blurry or shadowed images and eliminating duplicates captured from similar angles. Moreover, as most of the images in the dataset were taken from different patients or different parts of the same patient's body, they were treated as separate images. The pre-processed dataset contains a total of 572 images (Each image may contain multiple pin sites). To construct a pin site evaluation model, the location of the wound in the image needs to be extracted first. In our work, the skin around each pin on the image is included in the bounding box as a region of interest. In this step, we use yolov5\cite{annadatha2022preparing} to detect and extract the position of the pin, and then we intercept the bounding box as input. It is worth noting that only clearly visible pin is included in the bounding box, outliers such as pins that are difficult to identify due to obstructions, poor angles, or distance are ignored. The first row in Figure. \ref{fig_2} shows examples of pin site wound images from some patients. 

\subsection{Data annotation}
The dataset was labeled by two clinicians following the Modified Gordon Pin Infection Classification\cite{rahbek2021intrarater}. It is categorized into two distinct groups, namely ``Group A" and ``Group B". These two groups exhibit different characteristics, with ``Group A" images showing corresponding signs of inflammation/infection, and the MGS (Modified Gordon Score) of all images being greater than 0. In contrast, ``Group B" images show no obvious complications, with all of them having an MGS of 0. In cases where certain pins had different MGS, we employed the following approach: if both clinicians judged the images to be inflamed/infected (all of their MGS were higher than 0), these images were labeled as ``Group A". For other images where the MGS were not the same (e.g. one MGS score was 0 and the other was greater than 0), we excluded them from the dataset. It's important to emphasize that the clinicians labeled the results solely based on the visual appearance of the pin sites, without direct confirmation of their accuracy in reflecting the actual clinical conditions. The dataset consists of a total of 666 pin site photographs and contains two different categories: Group B (465) and Group A (201).

\subsection{Data augmentation}
We have three reasons for applying augmentation to the NCH dataset, which is listed as follows:

1.	The size of the dataset is too small, consisting of only 666 images, which is not sufficient to train a robust pin sites detection model with the expected accuracy.

2.	Due to the varying nature of real-world scenarios in which patients find themselves, such as uncontrollable lighting conditions, mobile phone pixels, various angles, distances, etc. The dataset cannot cover all scenes. Therefore, the model needs to be trained on a wide enough range of scenes to achieve generalization.

3.	To solve the overfitting problem.

In our work, data augmentation is achieved through 8 methods: horizontal flip, vertical flip, brightness change with an enhancement range between 1.1 and 1.5, contrast enhancement with an enhancement range between 1.1 and 1.5, rotation between -45°--45°, rotation after vertical, shearing between -16° and 16°, and Cutout (random deletion of a 30*30 rectangular area with a 0 fill). It is worth noting that each augmentation method was selected based on its relevance to real-world medical imaging scenarios\cite{rettenberger2023self}, ensuring that the augmented images remain reasonable and useful for training purposes. While more sophisticated augmentation strategies such as CutMix\cite{yun2019cutmix}, AutoAugment\cite{cubuk2018autoaugment}, and generative adversarial methods\cite{akrout2023diffusion} have the potential to further enhance model robustness. But these techniques introduce complexities in maintaining clinical relevance and authenticity of images. For example, creating synthetic images or employing algorithmically generated changes runs the risk of introducing non-existent pathology or anatomically impossible scenarios. In medical imaging, where accuracy is critical, such changes can lead to misinterpretations. Figure. \ref{fig_2} shows some examples of original photographs enhanced with these 8 image enhancement methods. 

In the end, the NCH dataset is initially divided into a training set and a test set in a 70:30 ratio. Within the training set, a further division creates training and validation subsets at an 80:20 ratio. Subsequently, data augmentation is subsequently applied to augment the training set's size, resulting in a total of 3,348 images.

\begin{figure}[!t]
\centering
\includegraphics[width=3.4in, height=3.5in]{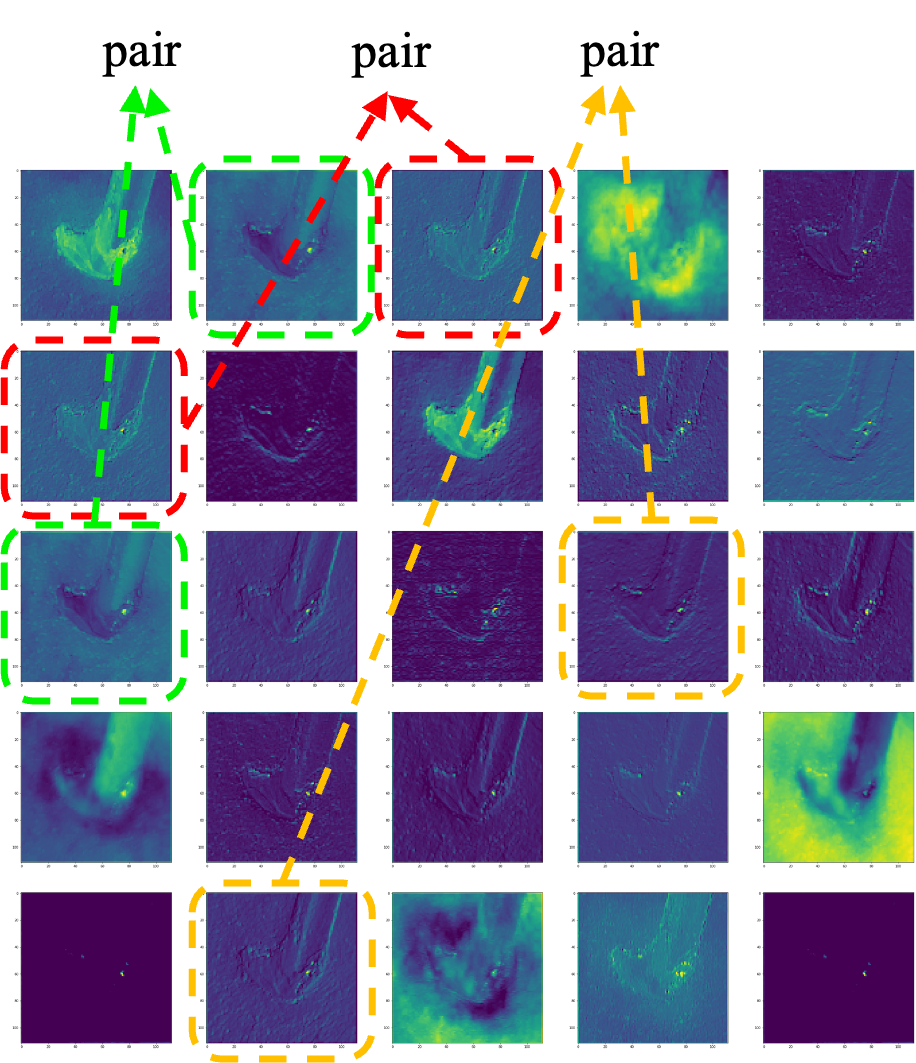}
\caption{Visualization of some similar redundant features generated by convolution in VGG. Feature redundancy can be reduced by cheap operational transformations.}
\centering
\label{fig_feature}
\end{figure}

\subsection{Network Architecture}
The pin site evaluation model consists of five blocks. The first two blocks consist of a convolutional layer (Conv), a batch normalization (BN) layer, and a corrected linear unit (ReLU). The last three blocks include inverted residuals block\cite{sandler2018mobilenetv2} with expansion dimension of 3, ERRC block, CBAM block\cite{woo2018cbam}, and ReLU6\cite{sandler2018mobilenetv2}. Figure. \ref{fig_3} illustrates the detailed structure of the wound classification model. The images are fed as input to the network to extract various visual features such as edges, texture, color, and angle. The resulting feature matrix (feature map) is used as the output. Then the output is flattened after all convolution operations have been performed and then sent to a fully connected dense layer to produce the final output. The final layer utilizes the ``softmax" activation function to classify the appearance of the images within our dataset.

To exclude the metal pins from the images, the model is focused on the skin surface of the wound by using the CBAM block. CBAM performs adaptive feature optimization by inferring the attention graph in two independent dimensions (channel and spatial) to improve the feature representation of the network. In addition, since we use ERRC to generate rich features, we need to use a channel attention mechanism to allow the network to automatically learn the importance of each channel feature map and adjust the feature response between channels accordingly. Trained convolutional networks on small datasets are often plagued with the problem of redundant information in their feature maps. To illustrate the problem, we randomly selected 25 channels from the output channels multiple times. Figure. \ref{fig_feature} displays these 25 feature maps. Upon observing the figure, it becomes evident that over 40\% of the feature maps are redundant, indicating that these feature maps convey similar information. This redundancy arises from the similarity of certain features across different channels of the CNN, leading to the extraction of similar features using the same convolutional kernel. 
To address this, we employ a cost-effective operation called ERRC to generate more diverse features and thereby minimize the redundancy in the information. Recent work\cite{chen2020dynamic}\cite{chen2019adaptive} has demonstrated that differently shaped convolutional kernels can come to handle the diversity of object features. To extract features effectively from small datasets, this method utilizes convolution kernels of varying sizes on half of the channels, leaving the other half unprocessed. By combining shallow and deep information through 1$*$1 convolution, the feature relationships between different channels can be explored more comprehensively, thus emphasizing the channels that play a key role in achieving accurate classification results. The whole process is as follows:

\begin{equation}
\label{errc}
\begin{split}
F = &K_{1\times 1}* Concat [I_{H\times W \times\frac{d}{4}}, I_{H\times W \times\frac{d}{4}},  \\
&(I_{H\times W \times\frac{d}{4}} *K_{3\times 3}), (I_{H\times W \times\frac{d}{4}}*K_{5\times 5})],
\end{split}
\end{equation}

\noindent where $F$ denotes the resultant feature produced by the above process, $I$ denotes the input feature, $K$ denotes the kernel whose subscript is the size of the kernel, and $Concat[\cdot ]$ denotes the splicing operation.

This approach enhances the expressiveness of the model, improves feature diversity, and effectively solves the feature redundancy problem. The implementation details of ERRC block are in Figure. \ref{fig_3}(c). It is worth noting that the number of parameters required for this approach is smaller than traditional convolution and inverted residuals block, which makes the whole network more lightweight.

\section{Experiments}
\subsection{Implementation details}
The equipment for all experiments was trained on an NVIDIA A4000 GPU with 16GB of RAM. The input image size is set to a uniform 224×224. In all experiments, Adam is used to optimize the model's objective, and a momentum of 0.92 is added to speed up convergence. The batch size is set to 16, and the maximum number of epochs is 300. The early stopping mechanism is used to halt training when the validation loss remains stable for 70 epochs. The learning rate is dynamically adjusted during training using continuous exponential decay, with an initial learning rate of 0.001, a decay period of 30, and a decay rate coefficient of 0.95. To prevent overfitting, we applied 0.2 and 0.2 dropout in the fully connected layer during training as regularization techniques. In addition, because there is a domain gap between the NCH dataset and some public data sets such as imagenet, we do not use any pre-training weights in training.

In this study, the NCH dataset we use is unbalanced. In reality, the number of true positives for a disease is much smaller than the number of true negatives. Specifically, the ratio of ``Group A" samples to ``Group B" samples in the dataset is 1:3.6, which could lead to a serious bias in our classification model. To mitigate this problem, we adopted Focal loss\cite{lin2017focal} as the loss function to allow the model to prioritize the learning of more challenging samples. By leveraging Focal loss, we were able to alleviate the class imbalance problem to some extent and improve the performance of the classification model.

\begin{equation}
\label{deqn_ex1a}
\begin{split}
FL_{loss} = &-(\alpha*(1-y_{p})^{\gamma}*y_{t}*\log_{}{y_{p}}+(1-\alpha)\\
&*(y_{p})^{\gamma}*(1-y_{t})*\log_{}{(1-y_{p})}),
\end{split}
\end{equation}

\noindent where $\alpha$ is the weight coefficient used for ``Group B" samples, and setting $\alpha$ achieves controlling the contribution of ``Group B" and ``Group A" samples to the loss. The $y_{t}$ denotes the true category of this sample and $y_{p}$ denotes the predicted category of this sample. $(1-y_{p})^{\gamma}$ is the adjustment factor, with tunable focusing parameter $\gamma$ $\ge$ 0. Its effect is to weaken the weight of sample losses that the model can already predict better. When ${\gamma}$ is 0, the loss function is equivalent to $\alpha$-balanced CE (cross entropy). In this paper, the value of $\alpha$ is 0.15, and the value of ${\gamma}$ is 2.

\begin{figure*}[!t]
\centering
\includegraphics[width=5.5in, height=2in]{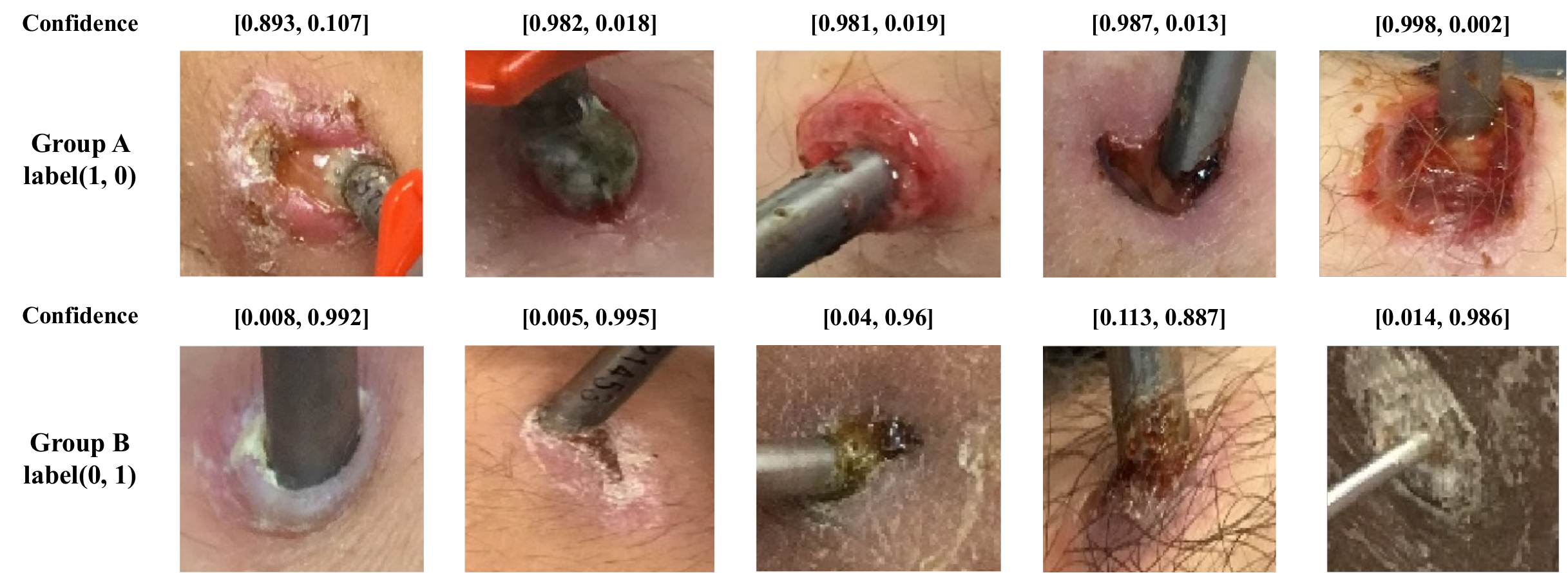}
\caption{Part of test results and confidence level of the model on the NCH dataset.}
\centering
\label{fig_4}
\end{figure*}

\subsection{Validation methods}
To evaluate the trained models, basic statistics such as true positives (TP), true negatives (TN), false positives (FP), and false negatives (FN) are calculated on the test dataset. These statistics represent the inflammation/infection detection rate, non-inflammation/uninfection detection rate, inflammation/infection false detection rate, and inflammation/infection missed detection rate, respectively. Accuracy, recall, and AUC (Area Under Curve) are used as evaluation metrics. Precision, which is the ability of a model to identify only relevant objects, is defined as the ratio of true positives to the sum of total positives and false positives. Recall (TPR), which is the ability of a model to find all relevant cases, is defined as the ratio of true positives to the sum of true positives and false negatives. The F1-score is a combination of precision and recall. The higher the F1-score, the more robust the classification model.

\begin{equation}
\label{deqn_ex2a}
Precision = \frac{TP}{TP+FP},
\end{equation}
\begin{equation}
\label{deqn_ex3a}
Recall = TPR = \frac{TP}{TP+FN},
\end{equation}
\begin{equation}
\label{deqn_ex4a}
F1 = 2*\frac{Precision* Recall}{Precision+Recall}.
\end{equation}

To ensure the efficiency of our model, we also compare the number of parameters in different modules, which corresponds to the space complexity of the module.
\begin{equation}
\label{conv}
Conv\ param = k^{2}*C_{in}*C_{out},
\end{equation}
\begin{equation}
\label{DSConv}\begin{split}
Inverted\ &residuals\ block\ param = a*C_{in}^{2}+ \\
&a*C_{in}*C_{out}+k^{2}*C_{in},
\end{split}
\end{equation}
\begin{equation}
\label{ERRC}
ERRC\ param = (3^{2}+5^{2})*\frac{C_{in}}{4}*\frac{C_{out}}{4}+C_{in}*C_{out},
\end{equation}

\noindent where $k$ is kernel size, $a$ is the expansion factor, $C_{in}$ denotes the number of the input channels, and $C_{out}$ refers to the number of the output channels. Given a kernel size of 3, input and output channels of 128 and 256 respectively. Compared to the convolution and inverted residuals block, the number of parameters of the ERRC module decreases by a factor of 2.88 and 1.42. This demonstrates that the model not only enhances feature richness but also conserves memory resources.

\section{Results and visualization}
To evaluate the effectiveness of the approach presented in this paper, we conducted testing of the network using the NCH dataset, subsequently obtaining training and detection results for the inflammation/infection task. In this section, we also performed a series of comparative experiments on the dataset to showcase the significance of each component of the proposed network structure, so as to better understand its detection performance and decision behavior. Additionally, we use the Gradient-weighted Class Activation Maps (Grad-CAM) technique to generate heatmaps of the classification outcomes, which aid in interpreting the model's decisions.

\begin{table}[!t]
\caption{Performance comparison of different baseline networks on the NCH dataset}
\label{alg:alg1}
\centering
\begin{tabular}{cccccc}
\hline
\hline
Network & Params & Prec\% & Recall\% & AUC & F1\\
\hline
VGG-16 & 14.77M & 99.4\% & 53.2\% & 0.807 & 0.69\\
\hline
VGG-19 & 20.08M & 98\% & 52.69\% & 0.822 & 0.685\\
\hline
ResNet-50 & 23.79M & 70\% & 80.46\% & 0.931 & 0.749\\
\hline
EfficientNetV2-b0 & 5.92M & 90\% & 88.23\% & 0.954 & 0.889\\
\hline
model without FL & 8.31M & 95.4\% & 79\% & 0.907 & 0.864\\
\hline
model with FL & 8.31M & 89.1\% & 90\% & 0.957 & 0.895\\
\hline
\bf{model with ERRC} & 5.77M & 92.5\% & 93.2\% & 0.975 & 0.927\\
\hline
\end{tabular}
\end{table}

\subsection{Results on NCH dataset}

To explore the potential of deep learning as a tool for distinguishing pin site types, we employed a trained model to predict the classification confidence of pin sites. Some of the classification results from our model are shown in Figure. \ref{fig_4}. The predicted outcome is a one-hot code that indicates whether the image belongs to ``Group A" or ``Group B". A threshold decision is then applied to assign a binary value, where 0 represents ``Group A" type and a value of 1 indicates ``Group B" type.

\begin{figure}[!t]
\centering
\includegraphics[width=3.3in,height=3in]{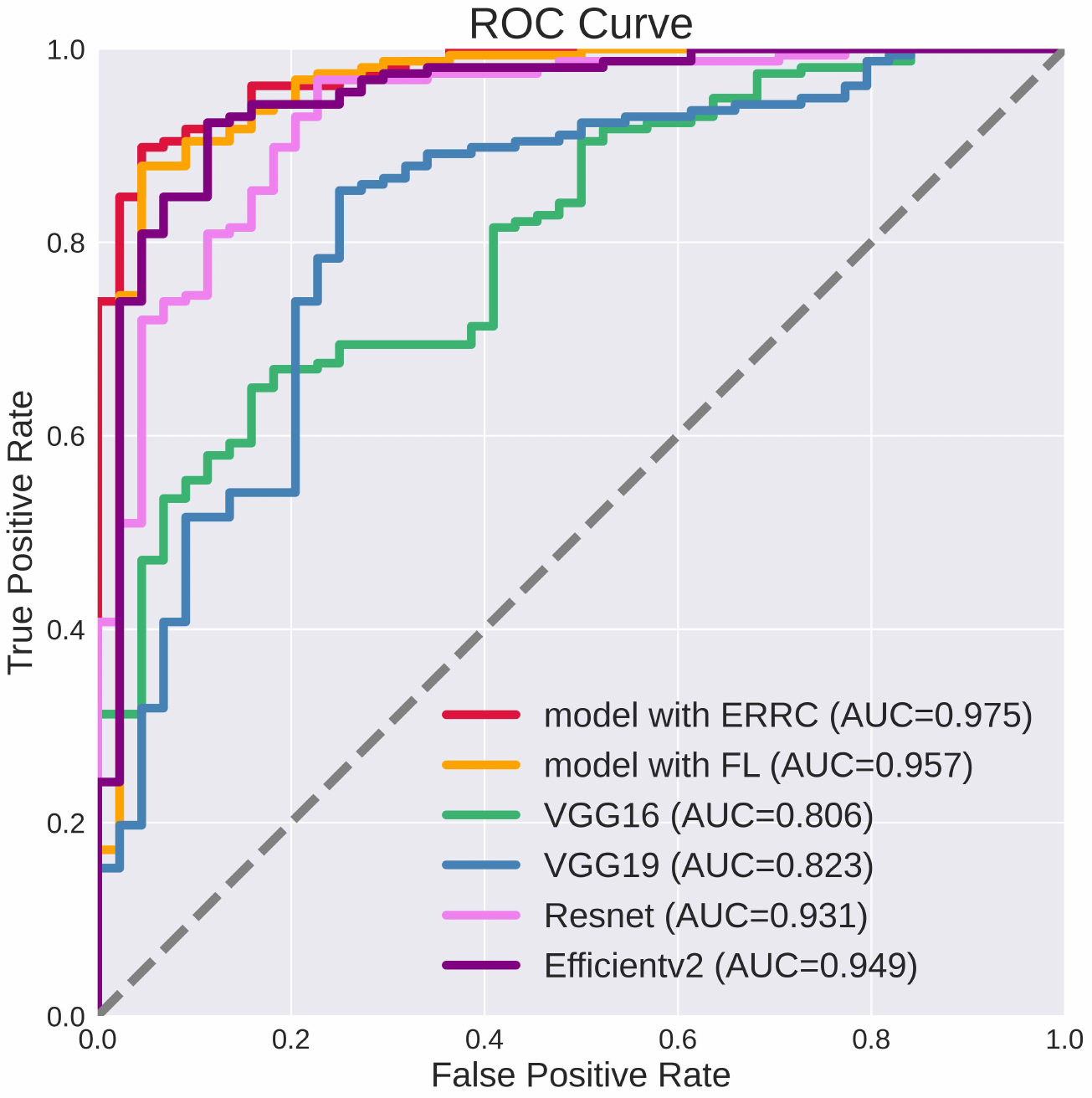}
\caption{ROC curves of different models on the test dataset.}
\label{fig_7}
\end{figure}

\begin{figure}[!t]
\centering
\includegraphics[width=3.4in, height=3.2in]{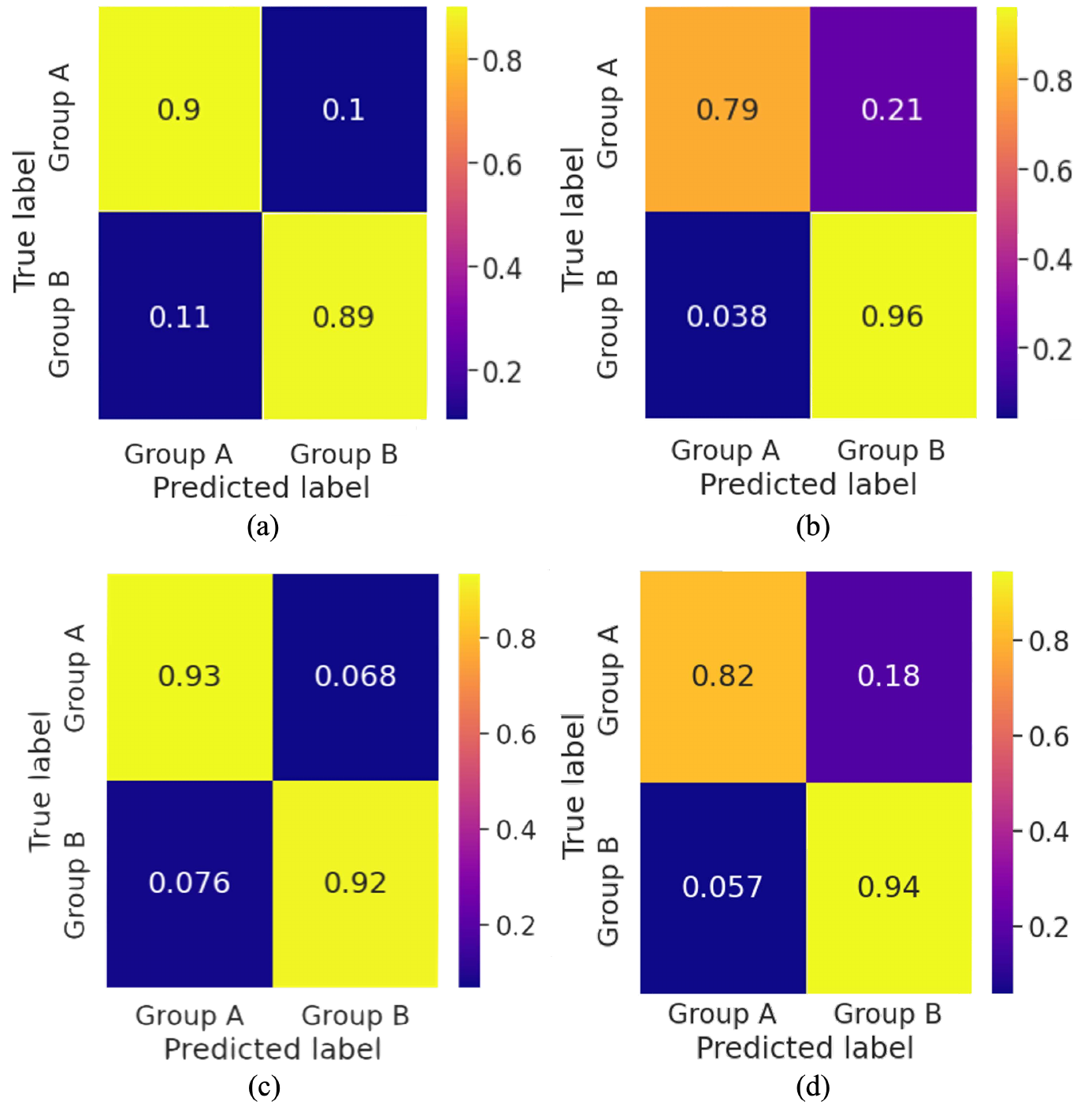}
\caption{Confusion matrix on the test dataset (a) model with FL, (b) model without FL, (c) model with ERRC and FL, (d) model with ERRC not FL.}
\centering
\label{fig6}
\end{figure}

We also compare our model with VGG16, VGG19\cite{simonyan2014very}, ResNet-50\cite{He_2016_CVPR}, and EfficientNetV2\cite{tan2021efficientnetv2}. Table \ref{alg:alg1} presents the evaluation results of the proposed model and commonly used classification models on the NCH dataset. We used the same data partitioning methods and evaluation metrics, including precision, recall, AUC, and F1-score, to train and evaluate the initialized network as well as the benchmark networks. Surprisingly, despite having the fewest parameters, our model outperformed these benchmark networks in terms of classification results. Our model achieved impressive results with a precision of 92.5\%, a recall of 93.2\%, and an F1 score of 0.927. The results of the dataset on the commonly used classification models are unsatisfactory. This discrepancy in results may be attributed to the dataset's limited size, which may not align with the design requirements of larger and deeper models. Additionally, the pin site image needs to be scaled up to a large size by interpolating the input image. As a result, this process can lead to a sparsity of useful information in the original images when they are enlarged to larger resolutions. Consequently, deeper models may tend to focus more on redundant information post-interpolation, potentially leading to performance degradation.

The results of changing the decision boundary are plotted on the ROC curve in Figure. \ref{fig_7}. As shown, the ROC curve of the proposed model is closest to the upper left corner, which indicates the best overall prediction performance of the model (AUC 0.975) and a significant advantage over several other models.

\subsection{Ablation study for modules}
To access the foundational performance of our model, we carried out an ablation study using the NCH dataset. In the evaluation of classification results, we utilized the confusion matrix to analyze the influence of incorporating focal loss into the model. Additionally, we conducted a comparative analysis to assess the impact of utilizing ERRC alongside inverted residual blocks within the model.

The prediction results of the model using the focal loss function and the cross-entropy loss function are shown in Figure. \ref{fig6}. A comparison of these results reveals that when employing the cross-entropy loss function, the model tends to focus more on the non-inflammation/uninfected category, achieving recognition rates of 96\% (inverted residuals block) and 94\% (ERRC) for ``Group B" categories. Although this is a high recognition rate, we should pay more attention to the accuracy of this category for this scenario that requires early prognosis, such as inflammated/infected. Table \ref{alg:alg1} presents the performance metrics from the experiments, including parameters, precision, recall, AUC, and F1-score. Based on the experiment results, it was observed that the precision of the models using the inverted residuals block and ERRC decreased by 6.3\% and 2.2\%, respectively, with the inclusion of focal loss. However, there was a significant increase in the recall of both models by 11\% and 13.5\%, respectively. These findings suggest that focal loss diminishes the impact of easy samples on the loss function, enabling the network to prioritize difficult samples. Consequently, this leads to improvements in the model's ability to identify "Group A" cases and enhances the feasibility of early prognosis. As a result, these models are better able to capture what is going on with the infection and minimize missed diagnoses.

\begin{figure}[] 
 \begin{center} 
  \vspace{-0.35cm}
  \subfigure[Feature maps using inverted residual block]
  { 
   \includegraphics[width=3in]{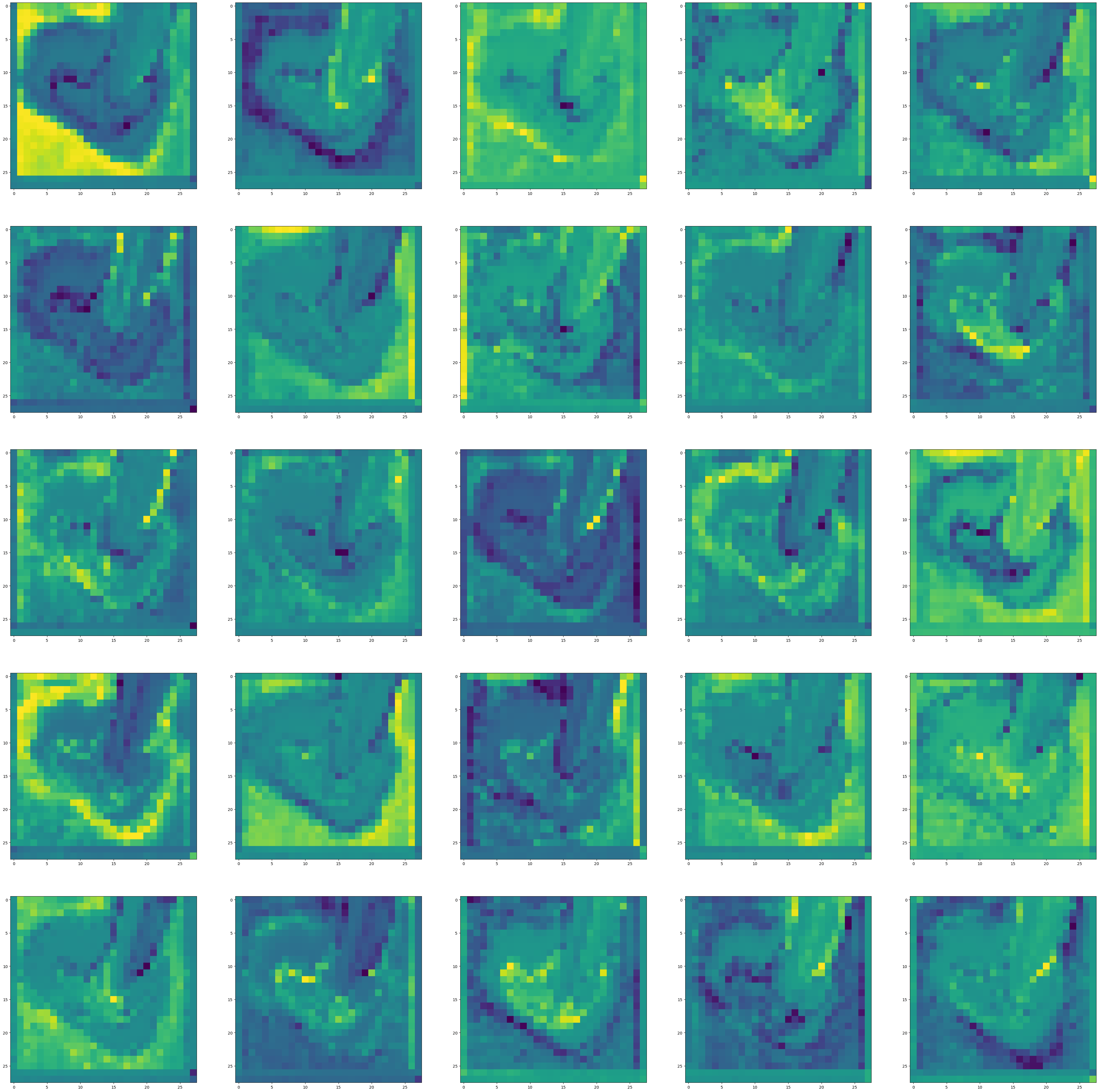} 
  }
  \subfigure[Feature maps using ERRC]
  { 
   \includegraphics[width=3in]{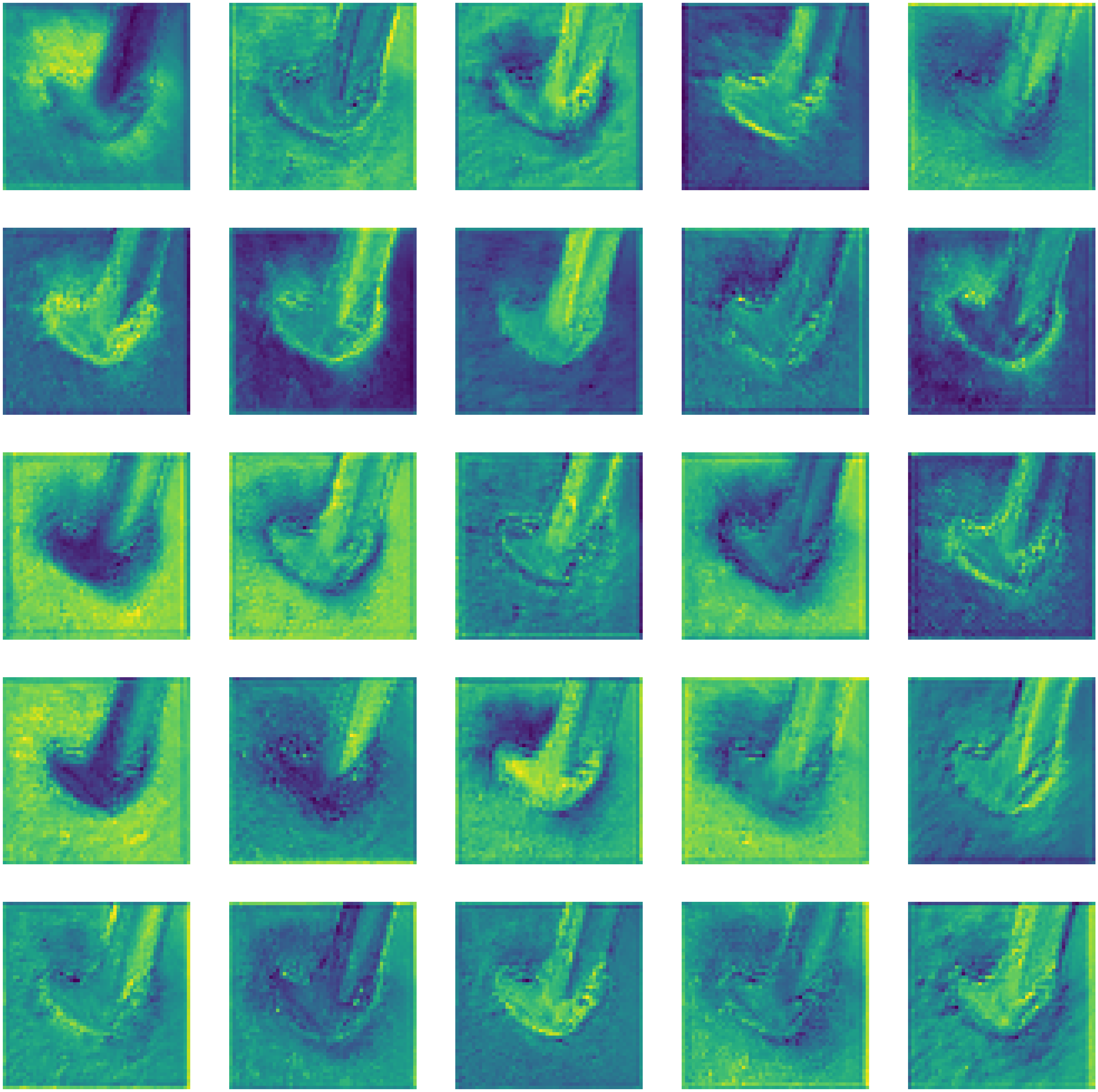} 
  } 
  \caption{Visualized comparison of the feature maps generated by the third block using different modules in the model.}
  \label{Fig.feature}
 \end{center} 
\end{figure}

With the simultaneous use of focal loss, compared to the inverted residual block, both precision and recall improved by 4.1\% and 2.5\% after using the ERRC module. We also noted enhancements in the AUC and F1-score of the model, which improved by 2.1\% and 3.2\%, respectively. In addition, Fig. \ref{Fig.feature} shows the feature maps obtained through the use of both the inverted residual block and the ERRC respectively. These feature maps are generated using a pin site image of the region of interest in Fig. \ref{fig_1}. Upon examining the feature maps generated by both methods, it becomes evident that the ERRC module outperforms in outlining the contours of the entire infected tissue. This finding demonstrates the enhanced capability of the ERRC module compared to the inverted residual block, specifically in extracting more intricate features using convolution kernels of various sizes. Significantly, this observation is crucial, as the delineated contour closely aligns with the boundary that separates normal skin from the wound tissue. This suggests that the effusive area of the wound contributes most to the judgment of wound infection, rather than the redness from scarring around the pin site, aligning with the Modified Gordon Pin Infection Classification. Therefore, we can assume that ERRC can improve the recognition rate of all categories by extracting richer features from images through the redundancy of the feature maps in the network. Moreover, the number of parameters of the model decreases by 30.6\% compared to the use of inverted residual block, which means that the performance of the model improves with a reduced number of parameters. However, the number of model parameters does not accurately reflect the model's inference speed. To address this, we conducted tests to measure the inference time and FPS of the models. We first selected models with an AUC value exceeding 0.9 to ensure high performance. In addition, to ensure the accuracy of the inference time measurements, we started the model with random inputs during the warm-up phase. After that, we performed 1000 tests on the device and calculated the average inference time and standard deviation, both of which are listed in Table 2. We can see that our proposed method achieves the lowest inference time and the highest FPS, which further proves the effectiveness and lightness of our model.

\begin{table}[!t]
\caption{Inference time and FPS for different models (batch size=1)}
\label{alg:alg2}
\centering
\resizebox{\linewidth}{!}{
\begin{tabular}{ccccc}
\hline
\hline
Network & Params & Mean time(ms) & Std time(ms) & FPS\\
\hline
ResNet-50 & 23.79M & 2.91 & 0.32 & 34.3\\
\hline
EfficientNetV2 & 5.92M & 3.12 & 0.39 & 32\\
\hline
model with FL & 8.31M & 2.48 & 0.09 & 40.4\\
\hline
\bf{model with ERRC} & 5.77M & 2.37 & 0.22 & 42.4\\
\hline
\end{tabular}}
\end{table}

\begin{figure*}[!t]
\centering
\includegraphics[width=7in]{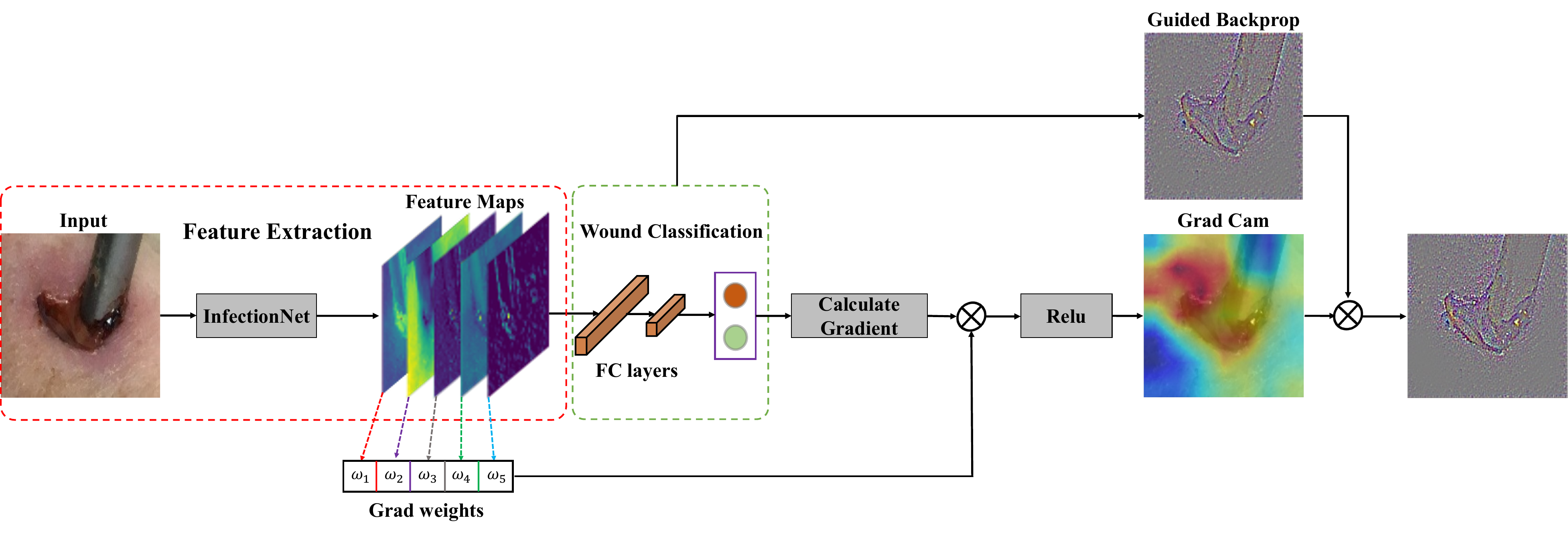}
\caption{Grad-CAM overview: given an image and a label as input, we multiply the heat map with the guided backpropagation to obtain the guided Grad-CAM visualization.}
\centering
\label{fig_8}
\end{figure*}

\begin{figure*}[!t]
\centering
\includegraphics[width=6in]{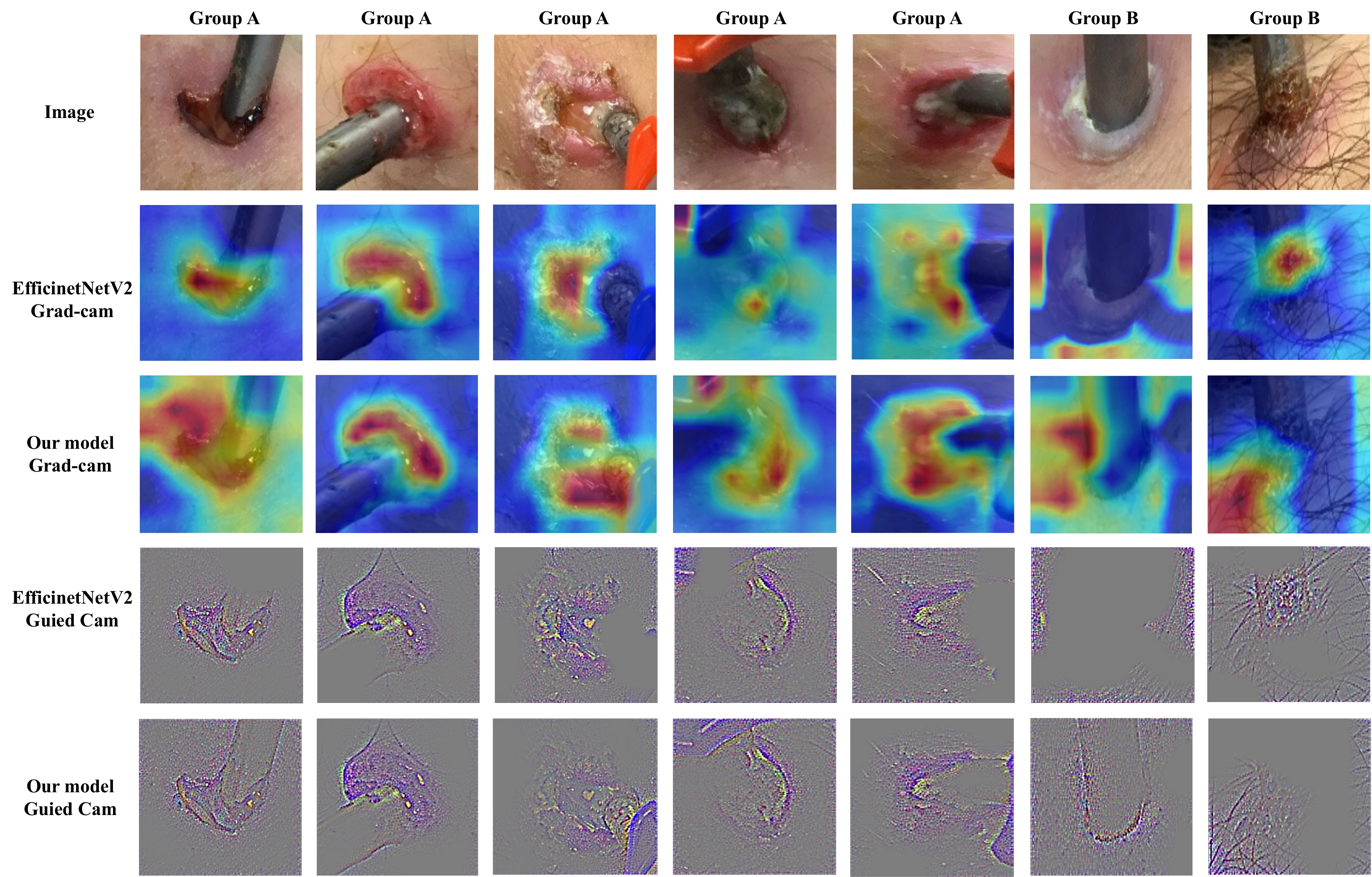}
\caption{The heatmap and Guided Cam results for our model and EfficientNetV2.}
\centering
\label{fig_9}
\end{figure*}

\subsection{Visualization of category activation in the attention module}
We utilized Grad-CAM, a technical tool for visualizing and explaining the decisions of the model, to gain a better understanding of how the network extracts features\cite{selvaraju2017grad}. When the model predicts the category at the output layer, the tool generates a heatmap of pixels based on the significance of each feature map, providing a visual representation of the model's attention. The regions to which the model allocates the highest attention are highlighted in dark red, while those receiving the least attention are depicted in dark blue. Figure. \ref{fig_8} illustrates the entire process of processing images using Grad-Cam to obtain interpretable images.

In the attention visualization process, we input an image along with its corresponding label, and employ back-propagation to compute the gradient of the feature values with respect to the input image. This helps us identify which parts of the image are most important in activating a particular feature. With the assistance of guided back-propagation, we can multiply the heatmap to achieve a guided Grad-CAM visualization that highlights the critical regions of the image that determine the state of the pin site. This process also enables us to visualize the results with high granularity at the pixel level. Figure. \ref{fig_9} depicts the visualization outputs between the proposed model and the baseline model on the dataset. Firstly, the area of interest of our model is expanded. Furthermore, the incorporation of CBAM also allows our model to effectively eliminate interference from the pin site, resulting in more precise localization and coverage of the skin tissue around the wound, aligning better with the physician's assessment criteria. Even when faced with interference from body hair, our model demonstrates the capability to mitigate attention spreading and bias, as illustrated in the last column in Figure. \ref{fig_9}. This observation highlights the network's ability to effectively extract relevant features from the skin tissue around the pin site and use them to classify the area effectively.

\begin{figure}[] 
 \begin{center} 
  \vspace{-0.35cm}
  \subfigure[training dataset result]
  { 
   \includegraphics[width=3in]{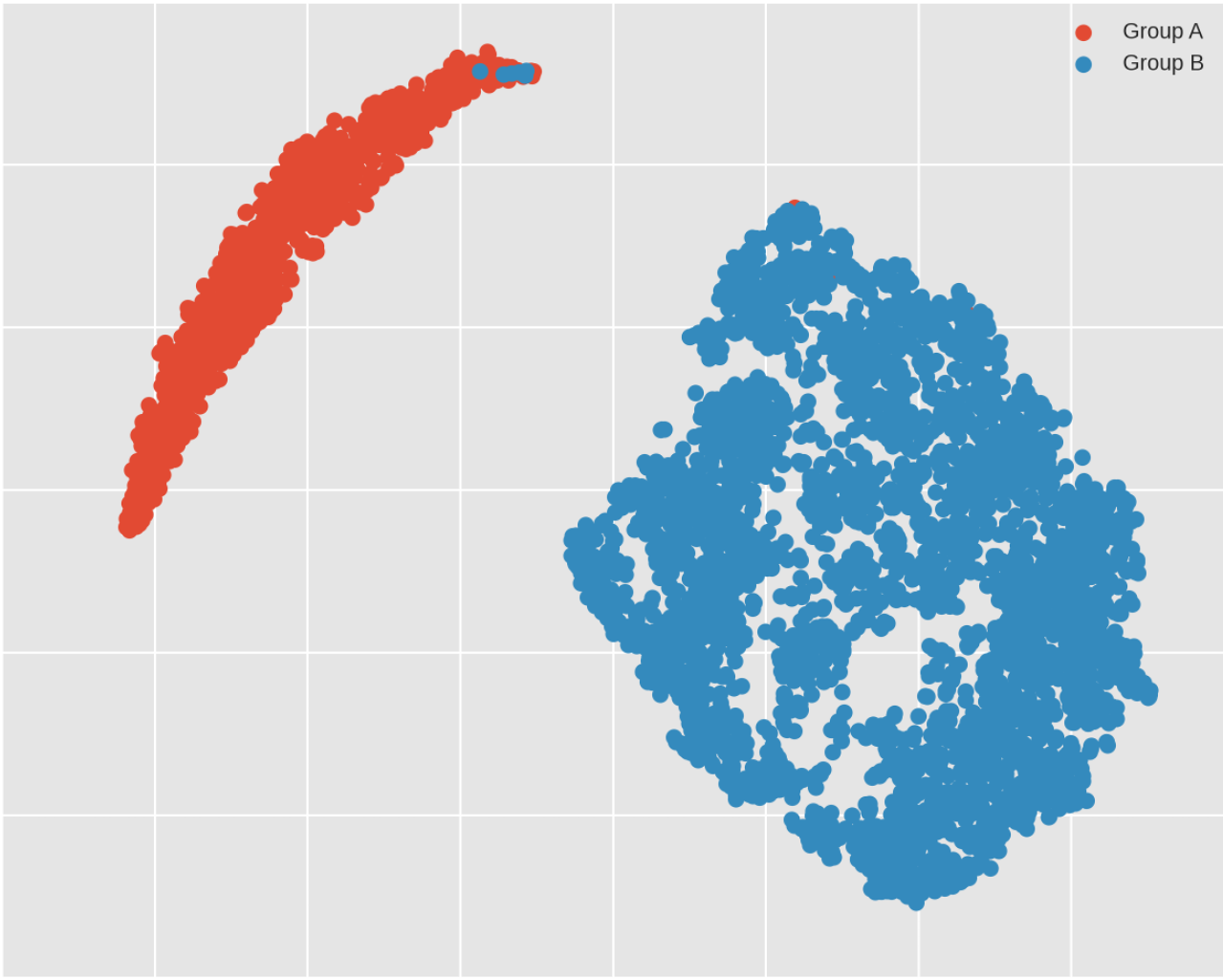} 
  }
  \subfigure[test dataset result]
  { 
   \includegraphics[width=3in]{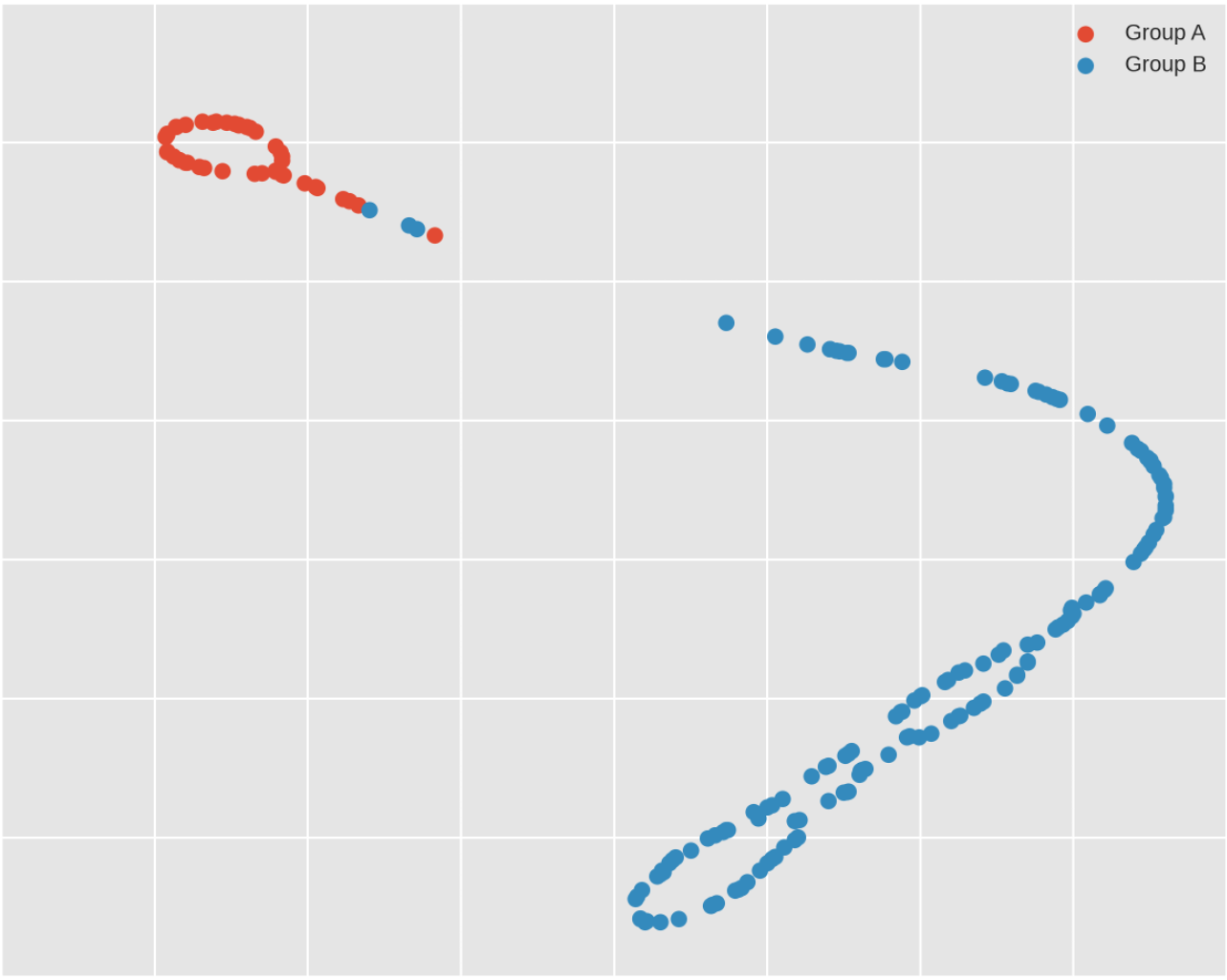} 
  } 
  \caption{2D visualization results of t-SNE on training dataset and test dataset.}
  \label{Fig.t-SNE}
 \end{center} 
\end{figure}

T-SNE (t-distributed Stochastic Neighbor Embedding) is an embedding model that transforms high-dimensional data into a low-dimensional space, while preserving local dataset characteristics\cite{van2008visualizing}. Visualizing high-dimensional data in a two-dimensional plot allows us to gain a better understand the model's classification performance for each data point.
The distribution of the model on the training and test sets is depicted in Figure. \ref{Fig.t-SNE}. In Figure. \ref{Fig.t-SNE}(a), the t-SNE dimensionality reduction mapping results on the training set are presented. It is apparent from the figure that the points belonging to ``Group A" and ``Group B" points are distinctly separated from each other, forming distinct clusters in different regions. Moving on to Figure. \ref{Fig.t-SNE}(b), we can observe the t-SNE results on the test set, which exhibit a similar pattern to the training set. This demonstrates that t-SNE effectively distinguishes the data in the test set based on the extracted features by the model. However, it is important to note that the test set may exhibit slight differences compared to the training set in terms of the mapping shape. These disparities may be due to variations in the data distribution caused by changes in patient cohorts or differences in the image acquisition environment. Additionally, the lower data density of the test set can also lead to slightly different results from the training set.

\section{Discussion}
We have proposed and evaluated a deep learning approach to investigate the model's ability and reliability in classifying different pin site statuses solely through photographs. Our preliminary results suggest that deep learning methods are reliable and accurate even with a limited training set size. By using the attention mechanism, the model can effectively mitigate pin site interference and classify wound states based on the skin's appearance near the pin sites. In addition, we have enhanced the representation of wound features using the ERRC module, which improves the classification ability and generalization of the network with reduced network parameters.

However, there is still much work to be done. Our study is limited by several factors: (1) We have not fully investigated the factors that may influence the results of pin site classification, such as differences in photographic equipment, the environment surrounding the wound (e.g. lighting), and internal factors (e.g. differences in patient skin color\cite{abubakar2019noninvasive}, use of antibiotic therapy or not). Due to the limited size of the dataset, it is not possible to draw conclusive evidence regarding the impact of skin color and antibiotic use on pin site appearance and healing. Further research is necessary to gain a deeper understanding of the influence of different factors on the test results. (2) For validation, we referred to the labels of medical experts, but these labels are based solely on the appearance of the wound and have not been verified for consistency with the actual inflammation/infection. Based on the good results so far, further validation and comparison with the results of medical tests could be the next step. (3) Only infected and non-infected conditions are considered, whereas the actual situation of the wound may be more complex. Future work could involve using methods such as few-shot learning and comparative learning to distinguish the degree of inflammation/infection in different wounds.

\section{Conclusion}
This paper explores the potential of deep learning in wound care by applying it to analyze wound images that include pin sites, with the aim of developing digital solutions that can replace routine hospital examinations. The classification models allow remote and non-contact wound assessment based on wound images, enabling early warning and intervention. To address the challenge of pin site interference and small datasets, this paper adopts an efficient CNN architecture with attention mechanism methods and redundancy reconstruction methods. The results show that the deep learning model effectively mitigates interference caused by pin sites in images and achieves a commendable pin site status classification with an AUC score of 0.975. These findings suggest that deep learning methods are a promising direction to explore for detecting the appearance of wounds and differentiating between them.

The main contributions of this paper are as follows.

1. A dataset was collected for lower limb pin site infections and an AI-based wound status classification model was developed that enables non-contact, low-cost wound assessment. Our results demonstrate the potential of deep learning to classify wound images based on visual features alone, highlighting the promising applications of this technology in wound care.

2. To address the characteristics of small datasets with limited data volume and redundant feature extraction, an efficient redundant reconstruction convolution has been developed to replace the inverted residual block. This reduces the number of model parameters while enriching features with less feature redundancy.

3. After generating a rich feature map using efficient redundant reconstruction convolution, the model enhances its recognition of key features through the application of channel and spatial attention convolutions. Additionally, it effectively excludes the interference of the pin sites in the image, focusing instead on extracting features from the skin tissue surrounding the wound.

4. A visual interpretation of the model's judgement results is offered, which is straightforward for physicians to comprehend and fosters further research possibilities. For example, it can be combined with medical identification results to discern the underlying pattern in the skin around wounds in different states (e.g. inflammation/infection).


\section*{References}
\def\refname{\vadjust{\vspace*{-2.5em}}} 

\bibliographystyle{IEEEtran}
\bibliography{ref}

\vspace{11pt}

\vfill

\end{document}